\documentclass[11pt,a4paper]{article}
\usepackage[hyperref]{emnlp2020}
\usepackage{times}
\usepackage{latexsym}

\usepackage[utf8]{inputenc}
\usepackage[T1]{fontenc}
\usepackage{hyphenat}
\usepackage{xspace}
\usepackage{amsmath}
\usepackage{amsfonts}
\usepackage{hyperref}
\usepackage{url}
\usepackage{booktabs}
\usepackage{multirow}
\usepackage{makecell}
\usepackage{caption}
\usepackage{minibox}
\usepackage{bbm}
\usepackage{graphicx}
\usepackage{balance}
\usepackage{mathtools}
\usepackage{color}
\usepackage{marvosym}
\usepackage{ifthen}
\usepackage{textcomp}
\usepackage{enumitem}
\usepackage{verbatim}
\usepackage{algorithm}
\usepackage{algorithmic}
\usepackage{numprint}
\usepackage{balance}

\usepackage{amsthm}
\theoremstyle{plain}

\newcommand{\chatoDisplayMode}[1]{#1}

\definecolor{MyRed}{rgb}{0.6,0.0,0.0} 
\definecolor{MyBlack}{rgb}{0.1,0.1,0.1} 
\newcommand{\inred}[1]{{\color{MyRed}\sf\textbf{\textsc{#1}}}}
\newcommand{\frameit}[2]{
  \begin{center}
  {\color{MyRed}
  \framebox[.9\columnwidth][l]{
    \begin{minipage}{.85\columnwidth}
    \inred{#1}: {\sf\color{MyBlack}#2}
    \end{minipage}
  }\\
  }
  \end{center}
}

\newcommand{\note}[2][]{\chatoDisplayMode{\def\@tmpsig{#1}\frameit{{\Pointinghand} Note}{#2\ifx \@tmpsig \@empty \else \mbox{ --\em #1}\fi}}}
\newcommand{\todo}[2][]{\chatoDisplayMode{\def\@tmpsig{#1}\frameit{{\Writinghand} To-do}{#2\ifx \@tmpsig \@empty \else \mbox{ --\em #1}\fi}}}

\newcommand{\abbrevStyle}[1]{#1}

\newcommand{\vs}{\abbrevStyle{vs.}\xspace}
\newcommand{\etc}{\abbrevStyle{etc.}\xspace}

\newcommand{\Secref}[1]{Sec.~\ref{#1}}
\newcommand{\Eqnref}[1]{Eq.~\ref{#1}}

\newcommand{\Tabref}[1]{Table~\ref{#1}}
\newcommand{\Figref}[1]{Fig.~\ref{#1}}

\newcommand{\Appref}[1]{Appendix~\ref{#1}}

\newcommand{\xhdr}[1]{\vspace{1.7mm}\noindent{{\bf #1.}}}

\newcommand{\textcite}[1]{\citeauthor{#1} \shortcite{#1}}

\newcommand{\cpt}[1]{\textsc{\MakeLowercase{#1}}}

\newcommand{\hide}[1]{}

\makeatletter
\newcommand{\iffont}[2]{\ifthenelse{\equal{\f@family}{#1}}{#2}{}}
\makeatother

  \usepackage{mathptmx}

  \DeclareSymbolFont{greek}{OML}{cmm}{m}{n}
  \DeclareMathSymbol{\alpha}{\mathalpha}{greek}{"0B}
  \DeclareMathSymbol{\beta}{\mathalpha}{greek}{"0C}
  \DeclareMathSymbol{\gamma}{\mathalpha}{greek}{"0D}
  \DeclareMathSymbol{\delta}{\mathalpha}{greek}{"0E}
  \DeclareMathSymbol{\epsilon}{\mathalpha}{greek}{"0F}
  \DeclareMathSymbol{\zeta}{\mathalpha}{greek}{"10}
  \DeclareMathSymbol{\eta}{\mathalpha}{greek}{"11}
  \DeclareMathSymbol{\theta}{\mathalpha}{greek}{"12}
  \DeclareMathSymbol{\iota}{\mathalpha}{greek}{"13}
  \DeclareMathSymbol{\kappa}{\mathalpha}{greek}{"14}
  \DeclareMathSymbol{\lambda}{\mathalpha}{greek}{"15}
  \DeclareMathSymbol{\mu}{\mathalpha}{greek}{"16}
  \DeclareMathSymbol{\nu}{\mathalpha}{greek}{"17}
  \DeclareMathSymbol{\xi}{\mathalpha}{greek}{"18}
  \DeclareMathSymbol{\pi}{\mathalpha}{greek}{"19}
  \DeclareMathSymbol{\rho}{\mathalpha}{greek}{"1A}
  \DeclareMathSymbol{\sigma}{\mathalpha}{greek}{"1B}
  \DeclareMathSymbol{\tau}{\mathalpha}{greek}{"1C}
  \DeclareMathSymbol{\upsilon}{\mathalpha}{greek}{"1D}
  \DeclareMathSymbol{\phi}{\mathalpha}{greek}{"1E}
  \DeclareMathSymbol{\chi}{\mathalpha}{greek}{"1F}
  \DeclareMathSymbol{\psi}{\mathalpha}{greek}{"20}
  \DeclareMathSymbol{\omega}{\mathalpha}{greek}{"21}
  \DeclareMathSymbol{\varepsilon}{\mathalpha}{greek}{"22}
  \DeclareMathSymbol{\vartheta}{\mathalpha}{greek}{"23}
  \DeclareMathSymbol{\varpi}{\mathalpha}{greek}{"24}
  \DeclareMathSymbol{\varrho}{\mathalpha}{greek}{"25}
  \DeclareMathSymbol{\varsigma}{\mathalpha}{greek}{"26}
  \DeclareMathSymbol{\varphi}{\mathalpha}{greek}{"27}
  \DeclareSymbolFont{otone}{OT1}{cmr}{m}{n}
  \DeclareMathSymbol{\Gamma}{\mathalpha}{otone}{0}
  \DeclareMathSymbol{\Delta}{\mathalpha}{otone}{1}
  \DeclareMathSymbol{\Theta}{\mathalpha}{otone}{2}
  \DeclareMathSymbol{\Lambda}{\mathalpha}{otone}{3}
  \DeclareMathSymbol{\Xi}{\mathalpha}{otone}{4}
  \DeclareMathSymbol{\Pi}{\mathalpha}{otone}{5}
  \DeclareMathSymbol{\Sigma}{\mathalpha}{otone}{6}
  \DeclareMathSymbol{\Upsilon}{\mathalpha}{otone}{7}
  \DeclareMathSymbol{\Phi}{\mathalpha}{otone}{8}
  \DeclareMathSymbol{\Psi}{\mathalpha}{otone}{9}
  \DeclareMathSymbol{\Omega}{\mathalpha}{otone}{10}
  \DeclareSymbolFont{syms}{OML}{cmm}{m}{it}
  \DeclareMathSymbol{\partial}{\mathord}{syms}{"40}
  \DeclareMathAlphabet{\mathbold}{OML}{cmm}{b}{it}
  \DeclareSymbolFont{largesymbols}{OMX}{cmex}{m}{n}

\usepackage{microtype}
\usepackage{multirow}
\usepackage[pass]{geometry}

\usepackage{graphicx}  
\usepackage{subcaption}

\usepackage{svg}

\usepackage{amsmath}
\usepackage{amssymb}
\usepackage{booktabs}
\usepackage{bbm}
\usepackage{amsfonts}       
\usepackage{nicefrac}       
\usepackage{microtype}      
\usepackage{amsthm}
\usepackage{thmtools, thm-restate}
\usepackage{natbib}
\usepackage{hyphenat}

\usepackage{mathtools}

\aclfinalcopy 

\title{KLearn: Background Knowledge Inference from Summarization Data}

\author{Maxime Peyrard \\
  EPFL \\
  \texttt{maxime.peyrard@epfl.ch} \\\And
  Robert West \\
  EPFL \\
  \texttt{robert.west@epfl.ch} \\}

\date{}

\begin{document}
\maketitle

\begin{abstract}
The goal of text summarization is to compress documents to the relevant information while excluding background information already known to the receiver.
So far, summarization researchers have given considerably more attention to relevance than to background knowledge.
In contrast, this work puts background knowledge in the foreground.
Building on the realization that the choices made by human summarizers and annotators contain implicit information about their background knowledge, we develop and compare techniques for inferring background knowledge from summarization data.
Based on this framework, we define summary scoring functions that explicitly model background knowledge, and show that these scoring functions fit human judgments significantly better than baselines.
We illustrate some of the many potential applications of our framework.
First, we provide insights into human information importance priors.
Second, we demonstrate that averaging the background knowledge of multiple, potentially biased annotators or corpora greatly improves summary\hyp scoring performance.
Finally, we discuss potential applications of our framework beyond summarization.
\end{abstract}




\section{Introduction}

Summarization is the process of identifying the most important information pieces in a document. For humans, this process is heavily guided by \emph{background knowledge}, which encompasses preconceptions about the task and priors about what kind of information is important \cite{Maybury:1999}.

\begin{figure}[ht]
  \centering
  \includegraphics[width=.8\columnwidth]{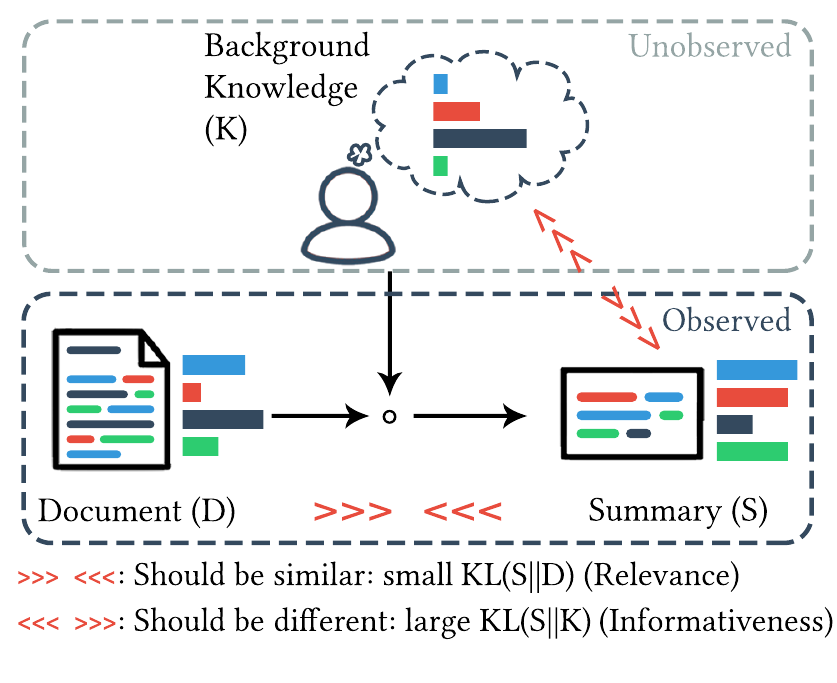}  
  \caption{A summary ($S$) results from the combination of the background knowledge ($K$) and the source document ($D$). Following \newcite{peyrard-2019-simple}, $S$ is similar to $D$ (\emph{Relevance} measured by a small $\cpt{KL}(S||D)$) but also brings new information compared to background knowledge (\emph{informativeness} measured by a large $\cpt{KL}(S||K)$).
We can infer the unobserved $K$ from the choices unexplained by the \emph{Relevance} criteria. }
  \label{fig:overall}
\end{figure}

    
    


Despite its fundamental role, background knowledge has received little attention from the summarization community. Existing approaches largely focus on the \emph{relevance} aspect, which enforces similarity between the generated summaries and the source documents \cite{peyrard-2019-simple}.

In previous work, background knowledge has usually been modeled by simple aggregation of large background corpora.
For instance, using \cpt{TF$\cdot$IDF} \cite{sparck1972statistical}, one may operationalize background knowledge as the set of words with a large document frequency in background corpora.

However, the assumption that frequently discussed topics reflect what is, on average, known does not necessarily hold. For example, common-sense information is often not even discussed \cite{conceptnet}. Also, information present in background texts has already gone through the importance filter of humans, e.g., writers and publishers. In general, a particular difficulty preventing the development of proper background knowledge models is its latent nature. We can only hope to infer it from proxy signals. Besides, there is, at present, no principled way to compare and evaluate background knowledge models.

In this work, we put the background knowledge in the foreground and propose to infer it from summarization data. Indeed, choices made by human summarizers and human annotators provide implicit information about their background knowledge. We build upon a recent theoretical model of information selection \cite{peyrard-2019-simple}, which postulates that information selected in the summary results from 3 desiderata: low \emph{redundancy} (the summary contain diverse information), high \emph{relevance} (the summary is representative of the document), and high \emph{informativeness} (the summary adds new information on top of the background knowledge). The tension between these 3 elements is encoded in a summary scoring function $\theta_K$ that explicitly depends on the background knowledge $K$.
As illustrated by \Figref{fig:overall}, the latent $K$ can then be inferred from the residual differences in information selection that are not explained by relevance and redundancy. For example, the black information unit in \Figref{fig:overall} is not selected in the summary despite being very prominent in the source document. Intuitively, this is explained if this unit is already \emph{known} by the receiver. 
To leverage this implicit signal, we view $K$ as a latent parameter learned to best fit the observed summarization data.

\xhdr{Contributions}
We develop algorithms for inferring $K$ in two settings: (i) when only pairs of documents and reference summaries pairs are observed (\Secref{ssec:without_h}) and (ii) when pairs of document and summaries are enriched with human judgments (\Secref{ssec:with_h}).

In \Secref{sec:comparison} we evaluate our inferred $K$s with respect to how well the induced scoring function $\theta_K$ correlates with human judgments. Our proposed algorithms significantly surpass previous baselines by large margins. 

In \Secref{sec:geometry}, we give a geometrical perpespective on the framework and show that a clear geometrical structure emerges from real summarization data.


The ability to infer interpretable importance priors in a data-driven way has many applications, some of which we explore in \Secref{sec:applications}. 
\Secref{sec:qualitative_analysis} qualitatively reveals which topics emerge as \emph{known} and \emph{unkown} in the fitted priors.
Moreover, we can infer $K$ based on different subsets of the data.
By training on the data of one annotator, we get a prior specific to this annotator. Similarly, one can find domain-specific $K$'s by training on different datasets. This is explored in \Secref{sec:annotator_specific}, where we analyze $16$ annotators and $15$ different summarization datasets, yielding interesting insights, e.g., averaging several, potentially biased, annotator-specific or domain-specific $K$'s results in systematic generalization gains.

Finally, we discuss future work and potential applications beyond summarization in \Secref{sec:ccl}. Our code is available at \url{https://github.com/epfl-dlab/KLearn}








\section{Related work}
The modeling of background knowledge has received little attention by the summarization community, although the problem of identifying content words was already encountered in some of the earliest work on summarization \cite{Luhn}.
A simple and effective solution came from the field of information retrieval, using techniques such as \cpt{TF}·\cpt{IDF} on background corpora  \cite{sparck1972statistical}. Similarly, \newcite{dunning1993accurate} proposed the log-likelihood ratio test to identify highly descriptive words. These techniques are known to be useful for news summarization \cite{Harabagiu:2005}. Later approaches include heuristics to identify summary-worthy bigrams \cite{bigram_worthy}. 
Also, \newcite{hong-nenkova-2014-improving} proposed a supervised model for predicting whether a word will appear in a summary or not (using a large set of features including global indicators from the New York Times corpus) which can then serve as a prior of word importance.

\newcite{conroy-schlesinger-oleary:2006} proposed to model background knowledge by aggregating a large random set of news articles. \newcite{Delort:2012} used Bayesian topic models to ensure the extraction of informative summaries. Finally, \newcite{louis:2014:P14-2} investigated background knowledge for update summarization with Bayesian surprise. 

These ideas have been generalized in an abstract model of importance \cite{peyrard-2019-simple} discussed in the next section.
\section{Background}
This work builds upon the abstract model introduced by \newcite{peyrard-2019-simple}, whose relevant aspects we briefly present here.

Let $T$ be a text and a function mapping a text to its semantic representation of the following form:
\begin{equation}
    \{\mathbb{P}_T(\omega_1), \dots, \mathbb{P}_T(\omega_n)\}
\end{equation}
The semantic representation is a probability distribution $\mathbb{P}$ over so-called \emph{semantic units} $\{\omega_j\}_{j \leq n}$. Many different text representation techniques can be chosen, e.g., topic models with topics as semantic units, or a properly renormalized semantic vector space with the dimensions as semantic units.

In the summarization setting, the source document $D$ and the summary $S$ are represented by probability distributions over the semantic units, $\mathbb{P}_D$ and $\mathbb{P}_S$.
Similarly, $K$, the background knowledge, is represented as a distribution $\mathbb{P}_K$ over semantic units.\footnote{We use $K$ and $\mathbb{P}_K$ interchangeably when there is no ambiguity.} Intuitively, $\mathbb{P}_K(\omega_j)$ is high whenever $\omega_j$ is \emph{known}. A summary scoring $\theta_K(S, D)$ (or simply $\theta_K(S)$ since the document $D$ is never ambiguous) can be derived from simple requirements:

\begin{align}
    \label{eq:eqn_theta}
    \theta_K(S) &= -\cpt{Red}(S) + \alpha \cdot \cpt{Rel}(S, D) + \beta \cdot \cpt{Inf}(S,K) \nonumber \\
    &= \cpt{H}(S) - \alpha \cdot \cpt{KL}(S\|D) +  \beta \cdot \cpt{KL}(S\|K),
\end{align}
where \cpt{Red} captures the redundancy in the summary via the entropy \cpt{H}. \cpt{Rel} reflects the relevance of the summary via the Kullback-Leibler (KL) divergence between the summary and the document. A \emph{good} summary is expected to be similar to the original document, i.e., the KL divergence $\cpt{KL}(S\|D)$ should be low. Finally, \cpt{Inf} models the informativeness of the summary via the KL divergence between the summary and the latent background knowledge $K$. The summary should bring new information, i.e., the KL divergence $\cpt{KL}(S\|K)$ should be high. 

In this work, we fix $\alpha=\beta=1$.

\section{The KLearn framework}
\label{sec:gen_framework}
As laid out, in our framework, texts are viewed as distributions over a choice of semantic units $\{\omega_j\}_{j \leq n}$. We aim to infer a general $K$ as the distribution over these units that best explains summarization data. We consider two types of data: with and without human judgments. 

\subsection{Inferring $K$ without human judgments}
\label{ssec:without_h}
Assume we have access to a dataset $\{x_i\}$ of pairs of documents $D_i$ and their associated summaries $S_i$: $x_i = (D_i, S_i)$. Under the assumption that the $S_i$ are \emph{good} summaries (e.g., generated by humans), we infer the background knowledge $K$ that best explains the observation of these summaries.  Indeed, if these summaries are \emph{good}, we assume that information has been selected to minimize \emph{redundancy}, maximize \emph{relevance} and maximize \emph{informativeness}. 

\xhdr{Direct score maximization}
A straightforward approach is to determine the $K$ that maximizes the $\theta_K$ score of the observed summaries. Formally, this corresponds to maximizing the function:
\begin{equation}
    \mathcal{F}_{MS}(K) = \left[\sum\limits_{x_i} \theta_{K}(x_i)\right] - \gamma \cdot \cpt{KL}(P\|K),
    \label{eq:naive_max}
\end{equation}
where $\cpt{KL}(P\|K)$ acts as a regularization term forcing $K$ to remain similar to a predefined distribution $P$.
Here, $P$ can serve as a prior about what $K$ should be. The factor $\gamma > 0$ controls the emphasis put on the regularization.


A first natural choice for the prior $P$ can be the uniform distribution $U$ over semantic units. In this case, we show in \Appref{sec:proofs} that maximizing \Eqnref{eq:naive_max} yields the following simple solution for $K$:
\begin{align}
    \mathbb{P}_K(\omega_j) \propto \sum\limits_{x_i=(D_i, S_i)} \left(\gamma - \mathbb{P}_{S_i}(\omega_j)\right).
\end{align}
With the choice $\gamma \geq 1$, note that $\mathbb{P}_K(\omega_j)$ is always positive, as expected.
This solution is fairly intuitive as it simply counts the prominence of each semantic unit in human-written summaries and considers the ones often selected as interesting, i.e., as having low values in the background knowledge. We denote this technique as \cpt{MS$|$U} to indicate the \emph{maximum score} with uniform prior. 
Surprisingly, it does not involve documents, whereas, intuitively, $K$ should be a function of both the summaries and documents. However, if such a simplistic model works well, it could be applied to broader scenarios where the documents may not even be fully observed.
 
Alternatively, we can choose the prior $P$ to be the source documents $\{D_i\}$. Then, as shown in \Appref{sec:proofs}, the solution becomes
\begin{align}
    \mathbb{P}_K(\omega_j) \propto \sum\limits_{x_i=(D_i, S_i)} \left(\gamma \cdot \mathbb{P}_{D_i}(\omega_j) - \mathbb{P}_{S_i}(\omega_j)\right).
\end{align}
Here a conservative choice for $\gamma$ to ensure the positivity of $\mathbb{P}_K(\omega_j)$ is $\gamma \geq \min\limits_j \frac{\mathbb{P}_S(\omega_j)}{\mathbb{P}_D(\omega_j)}$.
This model is also intuitive, as the resulting value of $\mathbb{P}_K(\omega_j)$ would be higher if $\omega_j$ is prominent in the document but not selected in the summary. This is, for example, the case for the black semantic unit in \Figref{fig:overall}. Furthermore, choosing $D$ as the prior implies viewing the documents as the only knowledge available and makes a minimal prior commitment as to what $K$ should be. We denote this approach as \cpt{MS$|$D}. 

\xhdr{Probabilistic model}
When directly maximizing the score of observed summaries, there is no guarantee that the scores of other, unobserved summaries remain low. A principled way to address this issue is to formulate a probabilistic model over the observations $x_i = (D_i,S_i)$:
\begin{equation}
    \mathbb{P}(x_i) = \frac{\exp(\theta_K(x_i))}{\sum\limits_{S \in Summ(D_i)} \exp(\theta_K((D_i, S))},
\end{equation}
where the partition function is computed over the set $Summ(D_i)$ of all possible summaries of document $D_i$. In practice, we draw random summaries as negative samples to estimate the partition function ($4$ negative samples for each positive). 

Then, $K$ is learned to maximize the log-likelihood of the data via gradient descent. To enforce the constraint of $K$ being a probability distribution, we parametrize $K$ as the softmax of a vector $\mathbf{k} = [k_1, \dots, k_n]$ of scalars. The vector $\mathbf{k}$ is trained with mini-batch gradient descent to minimize the negative log-likelihood of the observed data.
This approach is denoted as \cpt{PM}.

\subsection{Inferring $K$ with human judgments}
\label{ssec:with_h}
Next, we assume a dataset annotated with human judgments. Observations come in the form ${(S_i, D_i, h_i)}$ where $h_i$ is a human assessment of how good $S_i$ is as a summary of $D_i$. We can use this extra information to enforce high-scoring (low-scoring) summaries to also have a high (low) $\theta_K$ scores. 

\xhdr{Regression}
As a first solution, we propose regression, with the goal of minimizing the difference between the predicted $\theta_K$ and the corresponding human scores on the training set.
More formally, the task is to minimize the following loss:
\begin{equation}
    \mathcal{L}_{reg}(K) = \frac{1}{2} \sum_{x_i} (a \cdot \theta_K(x_i) - h_i)^2,
\end{equation}
where $a>0$ is a scaling parameter to put $\theta_K$ and $h_i$ on a comparable range. To train $K$ with gradient descent, we again parametrize $K$ as the softmax of a vector of scalars (cf. \Secref{ssec:without_h}). We denote this approach as \cpt{hReg}.

\xhdr{Preference learning}
In practice, regression suffers from annotation inconsistencies. In particular, the human scores for some documents might be on average higher than for other documents, which easily confuses the regression. Preference learning (PL) is robust to these issues, by learning the relative ordering induced by the human scores \cite{gao-etal-2018-april}. 
PL can be formulated as a binary classification task \cite{Maystre:255399}, where the input is a pair of data points $\{(S_i, D_i, h_i), (S_j, D_j, h_j)\}$ and the output is a binary flag indicating whether $S_i$ is better than $S_j$, i.e., $h_i > h_j$:
\begin{align}
    \mathcal{L}_{PL}(K) &= \sum\limits_{i,j} l(\sigma(\theta_K(x_i) - \theta_K(x_j)), \mathbbm{1}(h_i > h_j)),
\end{align}
where $\sigma$ is the logistic sigmoid function and $l$ can be, for example, the binary cross-entropy.
Again, we perform mini-batch gradient descent to train $\mathbf{k}$. We denote this approach as \cpt{hPL}.

\section{Comparison of approaches}
\label{sec:comparison}
To compare the usefulness of various $K$'s, we need a way to evaluate them. Fortunately, there is a natural evaluation setup: (i) plug $K$ into $\theta_K$, the summary scoring function described by \Eqnref{eq:eqn_theta}, (ii) use the induced $\theta_K$ to score summaries $S_i$, and (iii) compute the agreement with human scores~$h_i$. 

To distinguish between the algorithms introduced in \Secref{sec:gen_framework}, we adopt the following naming convention for scoring functions: if the background knowledge $K$ was computed using algorithm $A$, we denote the corresponding scoring function by $\theta_{A}$; e.g., $\theta_{\cpt{hPL}}$ is the scoring function where $K$ was inferred by \cpt{hPL}. 


\paragraph{Data.}
We use two datasets from the Text Analysis Conference (TAC) shared task: TAC-2008 and TAC-2009.\footnote{\url{http://tac.nist.gov/2009/Summarization/}, \url{http://tac.nist.gov/2008/}}
They contain 48 and 44 topics, respectively. Each topic was summarized by about $50$ systems and $4$ humans.
All system summaries and human-written summaries were manually evaluated by NIST assessors for readability, content selection with Pyramid \cite{nenkova-passonneau:2004}, and overall responsiveness \cite{Dang08overviewof, Dang09overviewof}.
In this evaluation, we focus on the Pyramid score, as the framework is built to model the content selection aspect.

\paragraph{Semantic units.}
As in previous work \cite{peyrard-2019-simple}, we use words as semantic units. In \Secref{sec:applications}, we also experiment with topic models. However, different choices of text representations can be easily plugged in the proposed methods. Words have the advantage of being simple and directly comparable to existing baselines.


\begin{table}[]
\centering
\resizebox{0.9\columnwidth}{!}{
\setlength{\tabcolsep}{15pt}
\begin{tabular}{@{}lcc@{}}
\toprule
    & Kendall's $\tau$ & MR \\
\midrule                            
\multicolumn{3}{l}{\emph{\textbf{Baselines}}} \\   
\hspace{6mm} \cpt{LR}                   &  .115      & 26.6    \\
\hspace{6mm} \cpt{ICSI}                 &  .139      & 25.3    \\
\hspace{6mm} $\cpt{KL}(S\|D)$                   &  .204      & 37.5    \\
\hspace{6mm} $\cpt{JS}(S\|D)$                   &  .225      & 35.7    \\
\hspace{6mm} $\theta_{\cpt{IDF}}$               &  .242      & 23.3    \\
\hspace{6mm} \cpt{$\theta_{\cpt{U}}$}           &  .202      & 23.8    \\
\midrule                            
\multicolumn{3}{l}{\emph{\textbf{Without human judgments (ours)}}} \\
\hspace{6mm} $\theta_{\cpt{MS|U}}$      &  .271      & 22.8  \\
\hspace{6mm} $\theta_{\cpt{MS|D}}$      &  \textbf{.295}      & \textbf{17.9}  \\
\hspace{6mm} $\theta_{\cpt{PM}}$        &  .269      & 19.8    \\
\midrule                            
\multicolumn{3}{l}{\emph{\textbf{With human judgments (ours)}}}   \\
\hspace{6mm} $\theta_{\cpt{hReg}}$      &  .227      & 21.8    \\
\hspace{6mm} $\theta_{\cpt{hPL}}$       &  \textbf{.285}      & \textbf{18.6}  \\
\midrule
\multicolumn{3}{l}{\emph{\textbf{Best training data fit}}}   \\
\hspace{6mm} Optimal ($\theta_\cpt{HPL}$)    &  .457      & 14.5  \\        
\bottomrule                            
\end{tabular}
}
\caption{Comparison of background knowledge based on: how well the induced $\theta_K$ correlates with humans (Kendall's $\tau$, higher is better) and how far human-written summaries are ranked compared to system summaries (MR, lower is better). The improvements of  $\theta_{\cpt{MS|D}}$ and $\theta_{\cpt{hPL}}$ over the baselines are significant (paired $t$-test, $p < 0.01$).}
\label{tab:eval-systems}
\end{table}

\paragraph{Baselines.}

For reference, we report the summary scoring functions of several baselines:
\textbf{LexRank} (\cpt{LR}) \cite{Erkan2004} is a graph-based approach whose summary scoring function is the average centrality of sentences in the summary. \textbf{\cpt{ICSI}} \cite{Gillick2009} scores summaries based on their coverage of frequent bigrams from the source documents. \textbf{$\cpt{KL}(S\|D)$} and \textbf{$\cpt{JS}(S\|D)$} \cite{Haghighi:2009} measure divergences between the distribution of words in the summary and in the sources. JS divergence is a symmetrized and smoothed version of KL divergence.
Additionally, we report the performance of choosing the uniform distribution for $K$ (denoted $\theta_{\cpt{U}}$) and an \cpt{IDF}-baseline where $K$ is built from the document frequency computed using the English Wikipedia (denoted as $\theta_{\cpt{IDF}}$).
For reference, we report the performance of training and evaluating $\theta_{\cpt{hPL}}$ on all data (denoted as \emph{Optimal}). This measures the ability of \cpt{HPL} to fit the training data.

\paragraph{Results.}
\Tabref{tab:eval-systems} reports the 4-fold cross-validation, averaged over all topics in both TAC-08 and TAC-09. The first column reports the Kendall's $\tau$ correlation between humans and the various summary scoring functions. The second column reports the mean rank (MR) of reference summaries among all summaries produced in the shared tasks, when ranked according to the summary scoring functions. Thus, lower MR is better.

First, note that even techniques that do not rely on human judgments can significantly outperform previous baselines. The results of $\theta_{\cpt{MS|D}}$ are particularly strong, with large improvements despite the simplicity of the algorithm. Indeed, $\theta_{\cpt{MS|U}}$ and $\theta_{\cpt{MS|D}}$ have a time complexity of $\mathcal{O}(n)$, where $n$ is the number of topics and run much faster than any other algorithm ($\approx 2$ seconds on a single CPU to infer $K$ from a TAC dataset). Despite being more principled, $\theta_{\cpt{PM}}$ does not outperform $\theta_{\cpt{MS|D}}$.


Improvements over baseline are also obtained by \cpt{hPL}, which leverages the fine-grained information of human judgments. However, even without benefiting from supervision, \cpt{MS$|$D} performs similarly to \cpt{hPL} without significant difference.
Also, as expected, the preference learning setup $\theta_{\cpt{hPL}}$ is stronger and more robust than the regression setup $\theta_{\cpt{hREG}}$, which does not significantly outperform the uniform baseline $\theta_{\cpt{U}}$.

Therefore, we use \cpt{hPL} when human judgments are available and \cpt{MS$|$D} when only document-summary pairs are available.

\begin{figure}
    \centering
    \includegraphics[width=0.9\columnwidth]{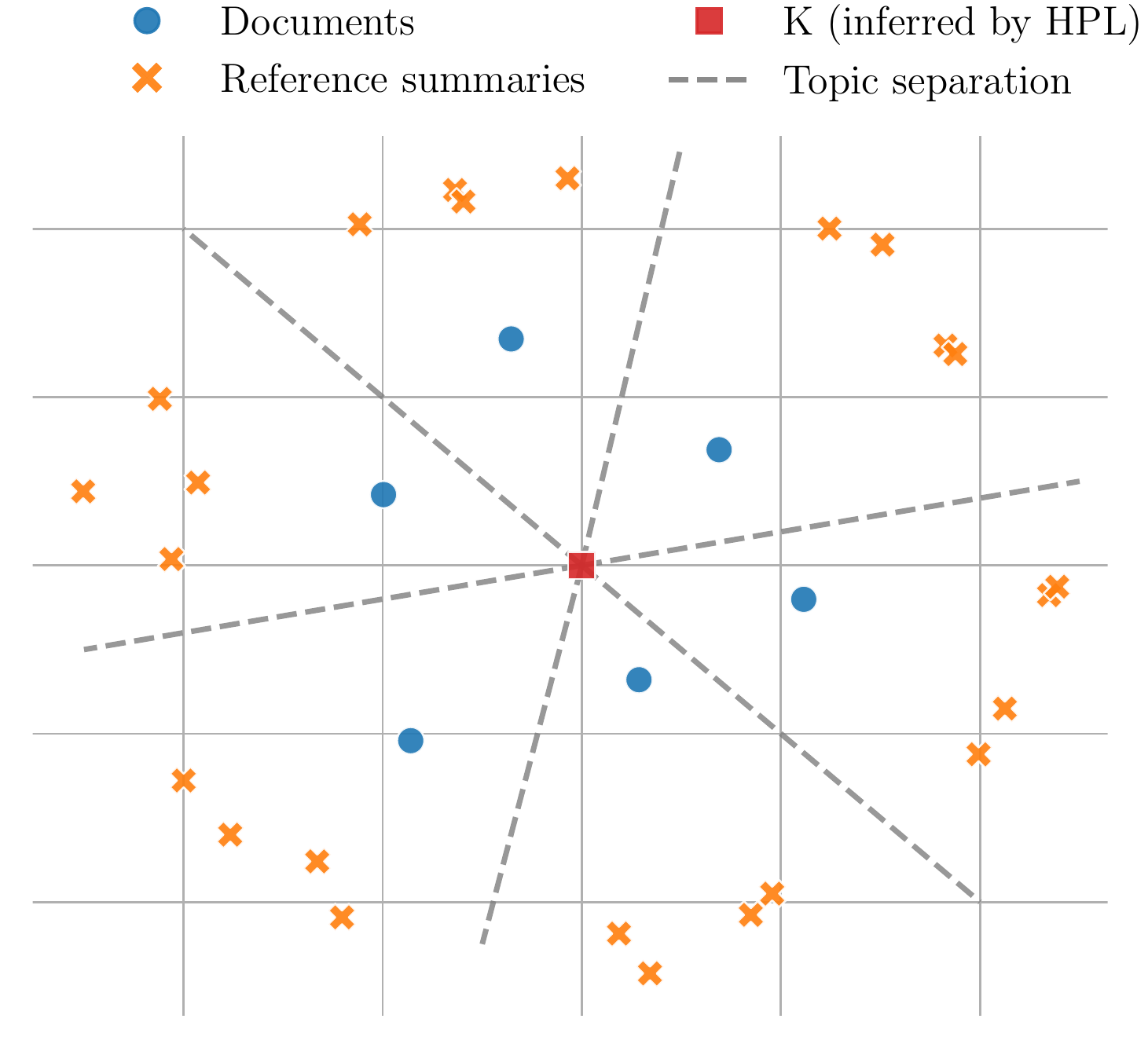}
    \caption{Multi-dimensional scaling projection of documents, summaries, and $K$ inferred by \cpt{hPL}. The Euclidean distance in the projection approximates to KL divergence in the original space. The geometrical intuition that summaries, documents, and $K$ should form a line with documents in the middle is simultaneously respected for 6 different randomly selected topics from TAC datasets.}
    \label{fig:geom}
\end{figure}

\section{A geometric view}
\label{sec:geometry}
Previously (see \Figref{fig:overall}), we mentioned that a good $K$ corresponds to a distribution such that the summary $S$ is different from $K$ ($\cpt{KL}(S\|K)$ is large) but still similar to the document $D$ ($\cpt{KL}(S\|D)$ is small). Furthermore, the regularization term in \Eqnref{eq:naive_max}, with $P=D$ enforcing small $\cpt{KL}(D\|K)$, makes minimal commitment as to what $K$ should look like, i.e., no a-priori information except the documents is assumed. 

Viewing these distributions as points in Euclidean space, the optimal arrangement for $S$, $D$, and $K$ is on a line with $D$ in between $S$ and $K$. 
Since human-written summaries $S$ and documents $D$ are given, inferring $K$ intuitively consists in discovering the point in high-dimensional space matching this property for all document-summary pairs. 

Interestingly, we can easily test whether this geometrical structure appears in real data with our inferred $K$. To do so, we perform a simultaneous multi-dimensional scaling (MDS) embedding of documents $D_i$, human-written summaries $S_i$, and $K$. In this space, two distributions are close to each other if their KL divergence is low. We plot such an embedding in \Figref{fig:geom} for 6 randomly chosen topics from TAC-09 and $K$ inferred by \cpt{hPL}. We indeed observe documents, summaries, and $K$ nicely aligned such that the summaries are close to their documents but far away from $K$. This finding also holds for $K$ inferred by \cpt{MS$|$D}.

These observations are important for two reasons.
(1)~They show that general framework introduced in \Figref{fig:overall} is an appropriate model of the summarization data:
For any given topic, the reference summaries are arranged on one side of the document. They deviate from the document in a systematic way that is explained by the repulsive action of the background knowledge. Human-written summaries contain information from the document but not from the background knowledge which puts them on the border of the space.
(2)~Our models can be seen to infer an appropriate background knowledge that is common to a wide spectrum of topics, as shown by the fact that $K$ occupies the central point in the embedding of \Figref{fig:geom}.

\section{Applications}
\label{sec:applications}
We now investigate some applications arising from our framework. As $K$ is easily interpretable, we explore which units receive high or low scores. One can also use different subsets (or aggregations) of training data. Here, we look into annotator-specific $K$'s and domain-specific $K$'s. 



\subsection{Qualitative analysis}
\label{sec:qualitative_analysis}
To understand what is considered as ``\emph{known}'' ($\mathbb{P}_K(\omega_j)$ is high) or ``\emph{unknown}'' ($\mathbb{P}_K(\omega_j)$ is low), we fit our best model, \cpt{hPL}, using all TAC data for two choices of semantic units: (i) words and (ii) LDA topics trained on the English Wikipedia (40 topics).


In \Tabref{tab:known_not_best_K} we report the top ``known'' and ``unknown'' words.
Frequent but uninformative words like `said' or `also' are considered known and thus undesired in the summary. On the contrary, unknown words are low-frequency, specific words that summarization systems systematically failed to extract although they were important according to humans. 
We emphasize that the inferred background knowledge encodes different information than a standard \cpt{IDF}. We provide a detailed comparison between $K$ and \cpt{IDF} in \Appref{sec:idfs_vs_optimal}.


\begin{table}
        \small
        \centering
        \begin{tabular}{@{}cc|cc@{}}
        \toprule
         \multicolumn{2}{c}{Known} & \multicolumn{2}{c}{Unknown}  \\
        \midrule
        said & say & kill & nation \\
        also & told & liberty & announcement \\
        like & one & new & investigation  \\
        \bottomrule
        \end{tabular}
        
        \caption{Example of words ``known'' and ``unknown'' according to the best $K$ inferred by \cpt{HPL}. A word $\omega_j$ is ``known'' (``unknown'') according to $K$ when $\mathbb{P}_K(\omega_j)$ is high (low).}\label{tab:known_not_best_K}
\end{table}

When using a text representation given by a topic model trained on Wikipedia, we obtain the following top 3 most known topics (described by 8 words):

\begin{enumerate}[topsep=0pt]
    \setlength\itemsep{-0.5em}
    \item government, election,  party, united, state, political, minister, president, \etc
    \item book, published, work, new, wrote, life, novel, well, \etc
    \item air, aircraft, ship, navy, army, service, training, flight, \etc
\end{enumerate}

The following are identified as the top 3 unknown topics:
\begin{enumerate}[topsep=0pt]
    \setlength\itemsep{-0.5em}
    \item series, show, episode, first, tv, film, season, appeared, \etc
    \item card, player, chess, game, played, hand, team, suit, \etc
    \item university, research, college, science, professor, research, degree, published, \etc
\end{enumerate}
Topics related to \emph{military} and \emph{politics} receive higher scores in $K$. Given that these topics tend to be the most frequent in news datasets, $K$ trained with human annotations learns to penalize systems overfitting on the frequency signal within source documents.
On the contrary, \emph{series}, \emph{games}, and \emph{university} topics receive low scores and should be extracted more often by systems to improve their agreement with humans.


\subsection{Inferring annotator- and domain\hyp specific background knowledge}
\label{sec:annotator_specific}
Within the TAC datasets, the annotations are also tagged with an annotator ID. It is thus possible to infer a background knowledge specific to each annotator, by applying our algorithms on the subset of annotations performed by the respective annotator. In TAC-08 and TAC-09 combined, $16$ annotators are identified, resulting in $16$ different $K$'s. 
\begin{figure}
    \centering
    \includegraphics[width=\columnwidth]{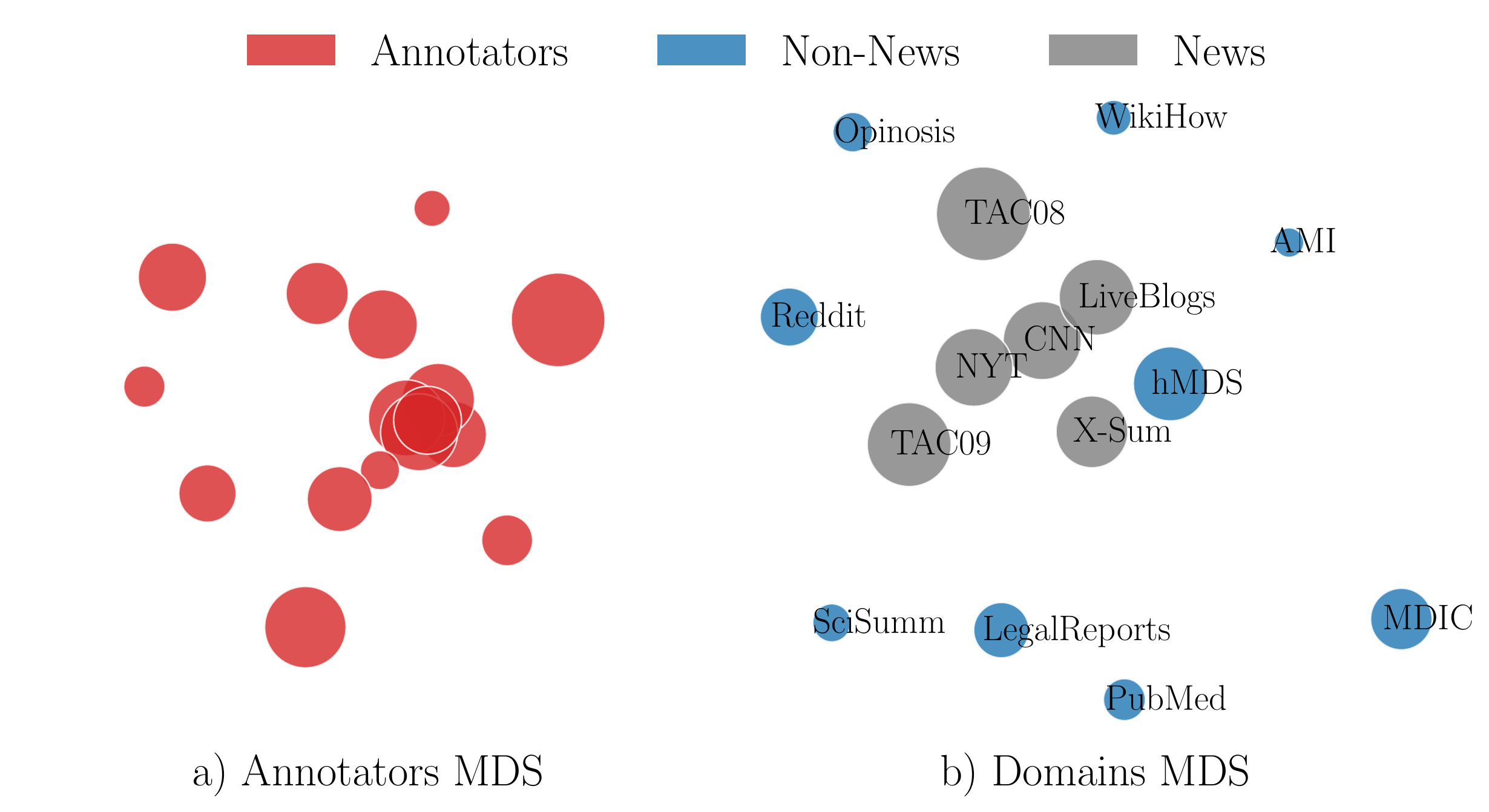}
    \caption{Multi-dimensional scaling projections of (a) annotators and (b) domains. The Euclidean distance in the projected space represents KL divergence in the original space. The disk size is proportional to how well the $K$ performs on the full TAC datasets, as evaluated by the correlation (Kendall's $\tau$) between the induced $\theta_K$ and human judgments.}
    \label{fig:mds}
\end{figure}

Instead of analyzing only news datasets with human annotations (like TAC), we can infer background knowledge from any summarization dataset from any domain as long as document--summary pairs are observed. To illustrate this, we consider a large collection of datasets covering domains such as news, legal documents, product reviews, Wikipedia articles, \etc
These do not contain human annotations, so we employ our \cpt{MS$|$D} algorithm to infer a $K$ specific to each dataset.
The detailed description of these datasets is given in \Appref{sec:datasets}.




\xhdr{Structure of differences}
To visualize the differences between annotators, we embed them in 2D using MDS with two annotators being close if their $K$ are similar. In \Figref{fig:mds} (a), each annotator is a dot whose size is proportional to how well its $K$ generalizes to the rest of the TAC datasets, as evaluated by the correlation (Kendall's $\tau$) between the induced $\theta_K$ and human judgments. The same procedure is applied to domains and is depicted in \Figref{fig:mds} (b).

News datasets appear at the center of all domains meaning that the news domain can be seen as an ``average'' of the peripheral non-news domains. Furthermore, the $K$'s trained on different news datasets are close to each other, indicating a good level of intra-domain transfer; and unsurprisingly, news datasets also exhibit the best transfer performance on TAC. 

\begin{figure}
    \centering
    \includegraphics[width=0.8\columnwidth]{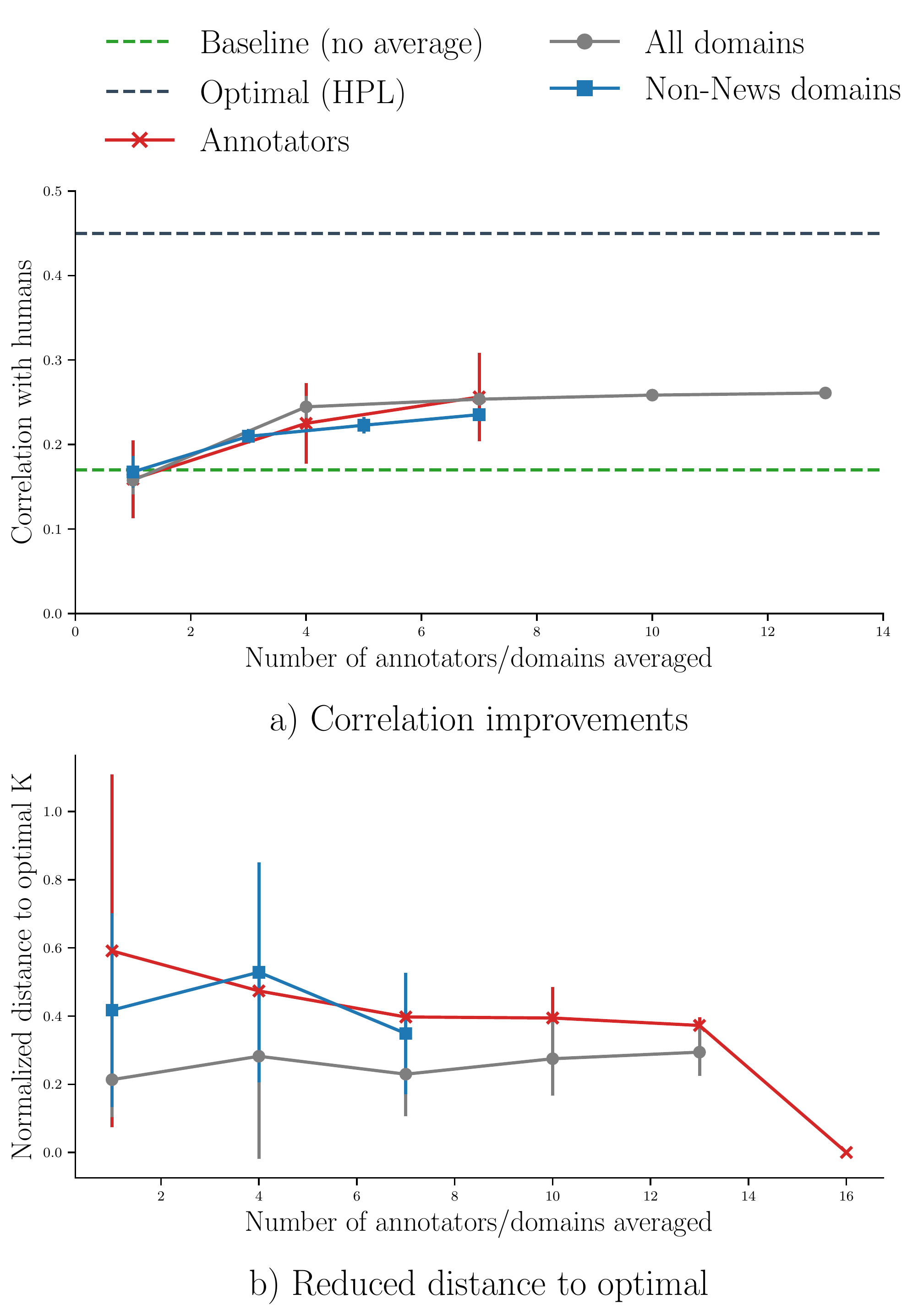}
    \caption{(a) Correlation with human judgments on TAC datasets (news domain) resulting from averaging annotator-specific $K$'s and domain-specific $K$'s. (b) Distance to the optimal $K$ (computed by running \cpt{HPL} on the full TAC datasets).}
    \label{fig:averaging}
\end{figure}

\begin{figure*}
    \centering
    \includegraphics[width=\textwidth]{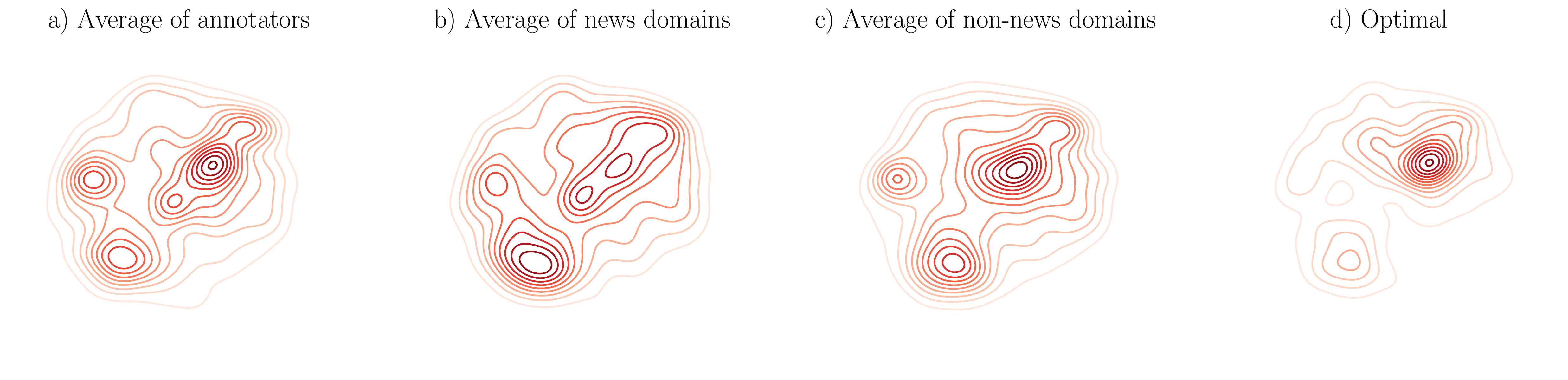}
    \caption{Several $K$ distributions visualized as density in 2D GloVe embedding space. }
    \label{fig:kde_plots}
\end{figure*}

\xhdr{Improvements due to averaging}
Based on previous observations, we make the hypothesis that averaging different annotator-specific $K$'s can lead to better correlation with human judgments on the unseen part of the TAC dataset. Similarly, news domains generalize better than other domains. We hypothesized that averaging domains may also result in improved correlations with humans in the news domain.

In \Figref{fig:averaging}(a), we report the improvements in correlation with human judgments on TAC (news domain) resulting from averaging an increasing number of annotators or domains. 
The error bars represent $95\%$ confidence intervals arising from selecting a different subset to compute the average. As we see, increasing the number of annotators averaged results in clear and significant improvements.
Since the error bars are small, which annotators are included in the averaging has little impact on the results.

Similarly, averaging different domains also results in significant improvements. In particular, averaging several non-news domains gives better generalization to the news domain.

Furthermore, \Figref{fig:kde_plots} shows, in the GloVe \cite{pennington-etal-2014-glove} embedding space, the $K$'s resulting from averaging (a) all annotators ($K$'s inferred by \cpt{hPL}), (b) all news datasets ($K$'s inferred by \cpt{MS$|$D}), and (c) all non-news datasets ($K$'s inferred by \cpt{MS$|$D}) in comparison to (d) the optimal $K$ learned with \cpt{hPL} trained on all data from TAC datasets. To produce these visualizations, we perform a density estimation of the $K$'s in the 2D projection of word embeddings. 

All averaged $K$'s tend to be similar to the optimal $K$. It indicates that only one prior produces strong results on the news datasets and it can be obtained by averaging many biased but different $K$'s. This is further confirmed by \Figref{fig:averaging}(b), where the distance to the optimal $K$ (measured in terms of KL divergence) significantly decreases when more annotators are averaged.\footnote{Note that the y-axis has been normalized to put the different divergences on a comparable scale.}

\section{Conclusion}
\label{sec:ccl}
We focus on the often-ignored background knowledge for summarization and infer it from implicit signals from human summarizers and annotators. We introduced and evaluated different approaches, observing strong abilities to fit the data.

The newly-gained ability to infer interpretable priors on importance in a data-driven way has many potential applications. For example, we can describe which topics should be extracted more frequently by systems to improve their agreement with humans. Using pretrained priors also helps systems to reduce overfitting on the frequency signal within source documents as illustrated by initial results in \Appref{sec:summarization}. 

An important application made possible by this framework is to infer $K$ on any meaningful subset of the data. In particular, we learned annotator-specific $K$'s, which yielded interesting insights: some annotators exhibit large differences from the others, and averaging several, potentially biased $K$'s results in generalization improvements. We also inferred $K$'s from different summarization datasets and also found increased performance on the news domain when averaging $K$'s from diverse domains.

For future work, different choices of semantic units can be explored, e.g., learning $K$ directly in the embedding space. Also, we fixed $\alpha=\beta=1$ to get comparable results across methods, but including them as learnable parameters could provide further performance boosts. Investigating how to infuse the fitted priors into summarization systems is another promising direction. 

More generally, inferring $K$ from a common-sense task like summarization can provide insights about general human importance priors. Inferring such priors has applications beyond summarization, as the framework can model any information selection task.

\section*{Acknowledgments}

We
gratefully
acknowledge partial support from
Facebook, Google, Microsoft,
the Swiss National Science Foundation (grant 200021\_185043),
and the European Union (grant 952215, ``TAILOR'').

\bibliography{KLearn_2020}
\bibliographystyle{acl_natbib}

\appendix

\clearpage

\section{2D Visualization of $K$}
\label{sec:viz}

\begin{figure}
    \centering
  \includegraphics[width=\linewidth]{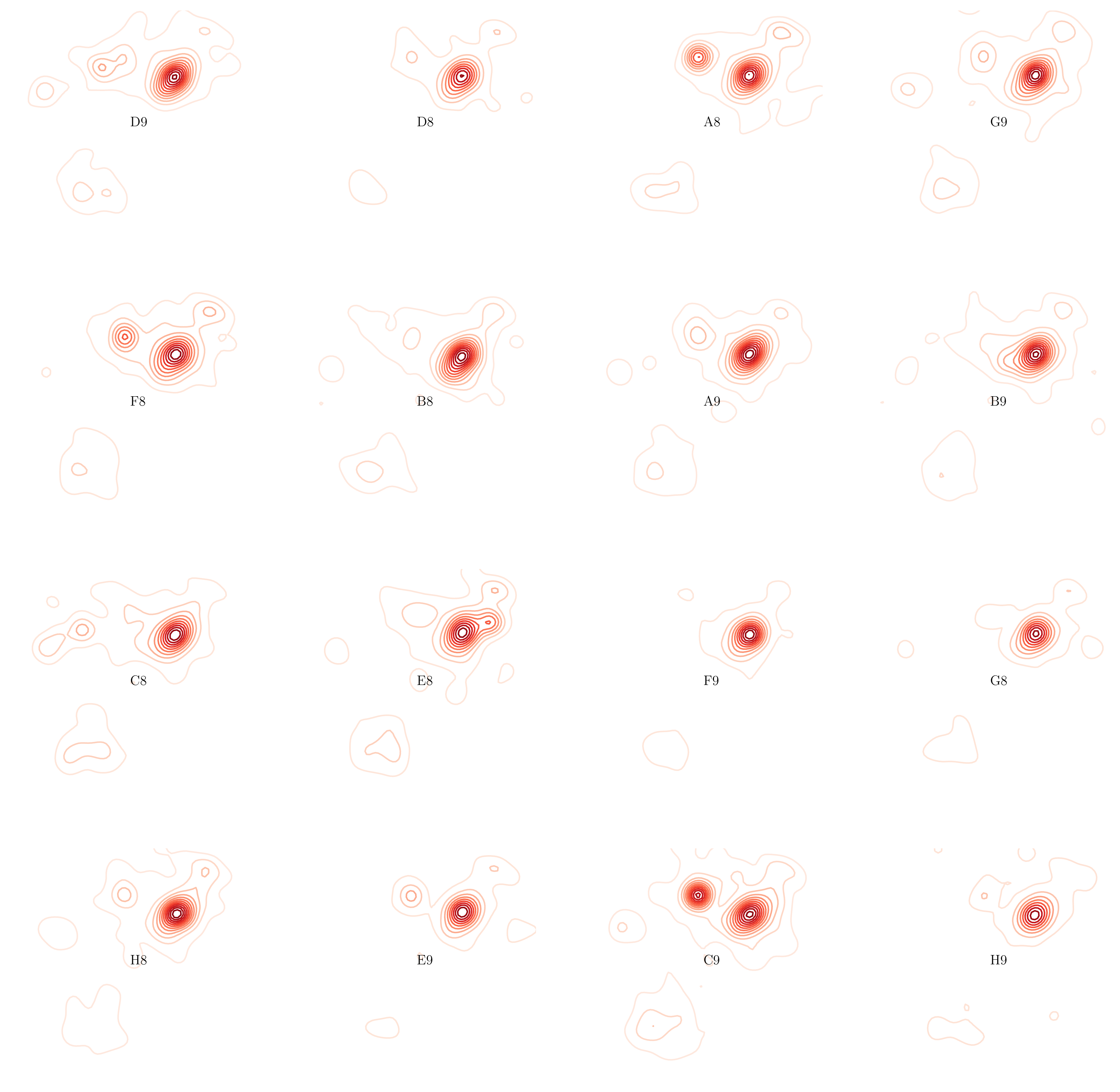}
  \caption{Visualization of each annotator's $K$ based on 2D projection of Glove word embedding.}
  \label{fig:annotator_viz}
\end{figure}

\begin{figure}
    \centering
  \includegraphics[width=\linewidth]{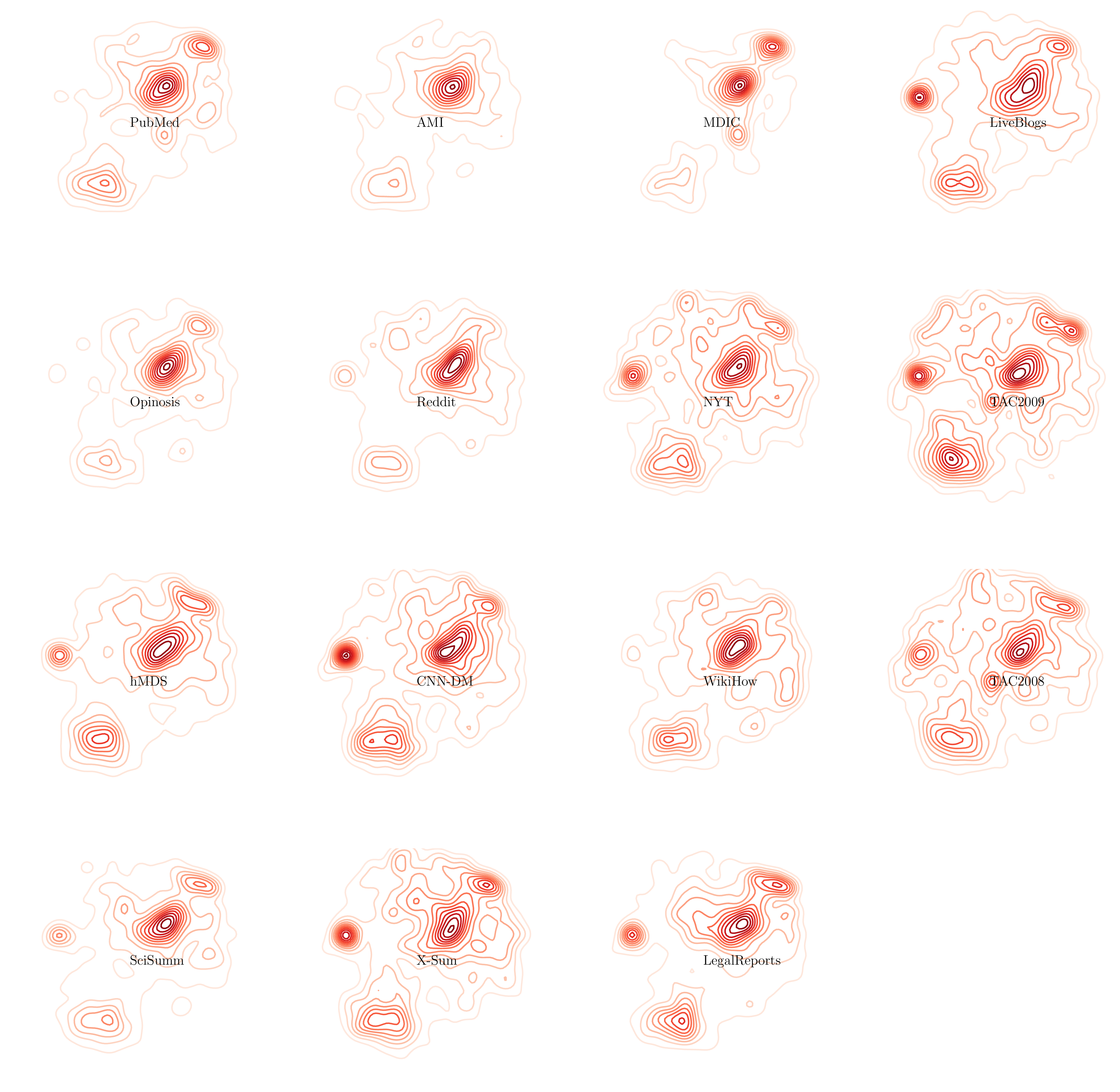}
  \caption{Visualization of each domain's $K$ based on 2D projection of Glove word embedding.}
  \label{fig:domain_viz}
\end{figure}

For each annotator and each domain, we produce visualizations in the 2D embedding space with the same procedure as in \Figref{fig:kde_plots}. 
\Figref{fig:annotator_viz} depicts the annotators and \Figref{fig:domain_viz} depicts the domains. It is interesting to observe much more diversity resulting from the domains and the domain-specific $K$'s are more spread out in the semantic space. This reflects the greater topic diversity discussed in different domains. In contrast, each annotator's $K$ is inferred based on the TAC datasets, which are in the same domain (news).

\section{Derivation of Approaches}
\label{sec:proofs}
The \textbf{direct score maximization} model consists in maximizing:
\begin{equation}
    \mathcal{L} = \sum\limits_{x} \theta_{K}(x) - \gamma \cdot \cpt{KL}(P||K),
\end{equation}

We use Lagrange multipliers with the constraint that $K$ is a valid distribution:
\begin{align}
    &L = \nonumber \\
    &\sum\limits_{x} \theta_{K}(x) - \gamma \cdot \cpt{KL}(P||K) - \lambda \left(\sum\limits_{\omega_j} \mathbb{P}_K(\omega_j) - 1\right)
\end{align}


(\cpt{MS$|$U}) First, with $P=U$ the uniform and $\gamma \neq 0$, we have the following derivatives:
\begin{align}
    &\frac{d \theta_K(x_i)}{d \mathbb{P}_K(\omega_j)} =  - \frac{\mathbb{P}_{S_i}(\omega_j)}{\mathbb{P}_K(\omega_j)} \\
    &\frac{d \gamma \cdot \cpt{KL}(U||K)}{d \mathbb{P}_K(\omega_j)} = - \frac{\gamma }{\mathbb{P}_K(\omega_j)} \\
    &\frac{d L}{d \mathbb{P}_K(\omega_j)} = -\sum\limits_{x}\frac{\mathbb{P}_{S_i}(\omega_j)}{\mathbb{P}_K(\omega_j)} + \frac{\gamma}{\mathbb{P}_K(\omega_j)} - \lambda
\end{align}
Setting the Lagrange derivative to $0$ yields:
\begin{align}
    \mathbb{P}_K(\omega_j) = \frac{1}{\lambda} \left(\gamma - \sum\limits_{x} \mathbb{P}_{S_i}(\omega_j)\right),
\end{align}
where $\lambda$ is the normalizing constant. In particular, when $\gamma = 1$:
\begin{align}
    \mathbb{P}_K(\omega_j) = \frac{1}{\lambda} \left(1 - \sum\limits_{x} \mathbb{P}_{S_i}(\omega_j)\right).
\end{align}
Note that choosing $\gamma \geq 1$ ensures that for all $\omega_j$, we have $\mathbb{P}_K(\omega_j) > 0$.

(\cpt{MS$|$D}) Second, we consider the case $P=D$ the document and $\gamma \neq 0$. $U$ changes with every document-summary pair and $\mathcal{L}$ becomes:
\begin{equation}
    \mathcal{L} = \sum\limits_{(D,S)} \theta(x) - \sum\limits_{(D,S)} \cpt{KL}(D||K),
\end{equation}
Then, only the the derivative concerning $\cpt{KL}(U||K)$ is modified and becomes:
\begin{equation}
    \frac{d \gamma \cdot \cpt{KL}(D||K)}{d \mathbb{P}_K(\omega_j)} = - \frac{\gamma \cdot \mathbb{P}_D(\omega_j)}{\mathbb{P}_K(\omega_j)}
\end{equation}
which gives the following solution after setting the Lagrange derivative to $0$:
\begin{align}
    \mathbb{P}_K(\omega_j) = \frac{1}{\lambda}\left(\sum\limits_{x} \gamma \cdot \mathbb{P}_{D_i}(\omega_j) - \mathbb{P}_{S_i}(\omega_j)\right).
\end{align}
Here it is not clear that $\mathbb{P}_K(\omega_j)$ is positive for every units. To avoid such issue, notice that we can choose $\gamma \geq \min_j \frac{\mathbb{P}_S(\omega_j)}{\mathbb{P}_D(\omega_j)}$.

\section{Datasets}
\label{sec:datasets}

The summarization track at the \emph{Text Analysis Conference} (TAC)  was a direct continuation of the DUC series. In particular, the main tasks of TAC-2008 \cite{Dang:2008overviewof} and TAC-2009 \cite{Dang:2009overviewof} were refinements of the pilot update summarization task of DUC 2007. A dataset of 48 topics was released as part of the 2008 edition and 44 new topics were created in 2009. TAC-2008 and TAC-2009 became standard benchmark datasets.

The \emph{New York Times Annotated Corpus} \cite{Sandhaus:2008} counts as one of the largest summarization datasets currently available. It contains nearly 1 million carefully selected articles from the \emph{New York Times}, each with summaries written by humans.

Also, the \emph{CNN/Daily Mail} dataset \cite{Hermann:2015:CNN-daily} has been decisive in the recent development of neural abstractive summarization \cite{see:2017,Paulus:2017,cheng-lapata:2016}. It contains \emph{CNN} and \emph{Daily Mail} articles together with bullet point summaries.

\newcite{Zopf:2016:hmds} also viewed the high-quality Wikipedia featured articles as summaries, for which potential sources were automatically searched on the web.

\cite{PVS:2018} recently crawled the live-blog archives from the \emph{BBC} and \emph{The Guardian} together with some bullet-point summaries reporting the main developments of the event covered.

To evaluate their opinion-oriented summarization system, \newcite{Ganesan:2010} constructed the \emph{Opinosis} dataset. It contains 51 articles discussing the features of commercial products (e.g., iPod's Battery Life).

Furthermore, we consider the large PubMed dataset \cite{cohan-etal-2018-discourse}, a collection of scientific publications.

The Reddit dataset \cite{kim-etal-2019-abstractive} has been collected on popular sub-reddits. 

The AMI corpus \cite{ami_corpus} is a standard product review summarization dataset.

\newcite{koupaee2018wikihow} automatically crawled the WikiHow website using the self-reported bullet points as summaries.

The XSUM dataset \cite{narayan-etal-2018-dont} is a large collection of news articles with a focus on abstractive summaries.

To measure the effect of information distortion in summarization cascades of scientific results, \newcite{mdic} collected manual summaries of various lengths.

We also included the LegalReport dataset \cite{legalreports} where the task is to summarize legal documents.

\begin{table*}
        \small
        \centering
        \begin{tabular}{@{}l|c|cc|c|cc@{}}
        \toprule
        Dataset     & Creation & \multicolumn{2}{c|}{Input}  & \multicolumn{1}{c|}{Summary} &  \multicolumn{2}{c}{Size}    \\
            & Man./Auto. & Type & Genre & Length &  Topics & Doc/Topic  \\
        \midrule
        TAC-2008         & M & MDS & News    & 100 &  48 & 10  \\
        TAC-2009         & M & MDS & News    & 100 &  44 & 10  \\
        SciSumm          & A & MDS & Sci.    & 150 &  1000 &  1 \\
        CNN/Daily Mail       & A & SDS & News   & $\approx$50 &  $\approx$300K & 1  \\
        NYT Corpus           & A & SDS & News   & $\approx$50 &  $\approx$650K & 1  \\
        Opinosis                    & M & MDS & Review    & $\approx$20 &  51 & $\approx$100  \\
        LiveBlogs         & A & Temporal & Snippets    & $\approx$60 & $\approx$2K & $\approx$70  \\
        hMDS                    & M & MDS & Heter.  & $\approx$216 &  91 &  $\approx$14 \\
        PUBMED                  & A & SDS & Sci.  & $\approx$100 & $\approx$133K & 1 \\
        XSUM                  & A & SDS & News    & $\approx$25 & $\approx$220K & 1 \\
        Reddit                  & A & SDS & Heter.    & $\approx$20 & $\approx$122K & 1 \\
        AMI                  & M & MDS & Meeting& $\approx$280 & 137 & 10 \\
        WikiHow                  & A & SDS & Heter.    & $\approx$60 & $\approx$230K & 1 \\
        LegalReports            & A & SDS & Legal   & $\approx$280 & 3500 & 1 \\
        MDIC                    & M & Cascade & Sci. & varying & 16 & 1 \\
        \bottomrule
        \end{tabular}
        \caption{Description of datasets used in the experiments}\label{tab:datasets_description}
\end{table*}

\section{Extracting Summaries: Example}
\label{sec:summarization}
Once $K$ is specified, the summary scoring function $\theta_K$ can be used to extract summaries. For extractive summarization, this is an optimal subset selection problem \cite{McDonald2007}.
Unfortunately, $\theta_K$ is not linear and cannot be optimized with Integer Linear Programming.
It is also not submodular and cannot be optimized with the greedy algorithm for submodularity.
We have to rely on generic optimization techniques which do not make any assumption about the objective function and can approximately optimize any arbitrary function.
We use the genetic algorithm proposed by \newcite{TUD-CS-20164649}\footnote{\url{https://github.com/UKPLab/coling2016-genetic-swarm-MDS}} which creates and iteratively optimizes summaries over time. 
We denote as ($\theta_K$, Gen) the summarization system approximately solving the subset selection problem. We compare $3$ systems: when $K$ is inferred by $\cpt{MS|D}$, when $K$ is inferred by \cpt{hPL} and when $K$ is the uniform distribution.\footnote{We employ a 3-fold cross-validation setup.}
For reference, we report the standard summarization baselines described in the previous section. The summaries are evaluated with 2 automatic evaluation metrics: ROUGE-2 recall with stopwords removed (R-2) \cite{Lin:2004} and a recent BERT-based evaluation metric (MOVER) \cite{zhao-etal-2019-moverscore}.
The results, reported in \Tabref{tab:summary_eval}, are encouraging since the systems based on the learned priors outperform the uniform prior. They also perform well in comparison to baselines. The inferred prior can benefit systems by preventing them from overfitting on the frequency signal.
\begin{figure}
    \centering
    \begin{subfigure}{\columnwidth}
    \includegraphics[width=0.9\columnwidth]{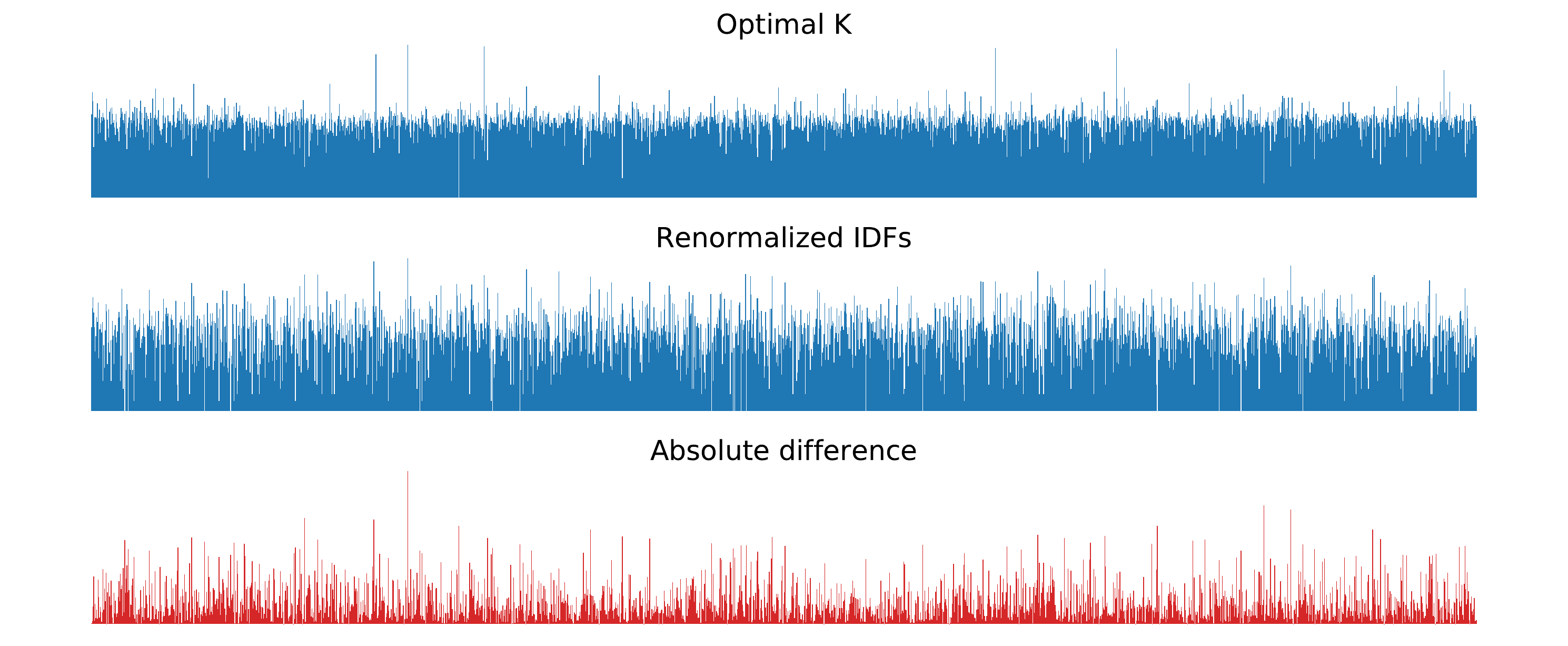}
    \caption{Visualization of optimal $K$, renormalized \cpt{IDF} and their absolute difference. One bar for word in the support ($2000$  words). }
    \label{fig:direct_comparison}
    \end{subfigure}
    
    \begin{subfigure}{\columnwidth}
    \includegraphics[width=0.9\columnwidth]{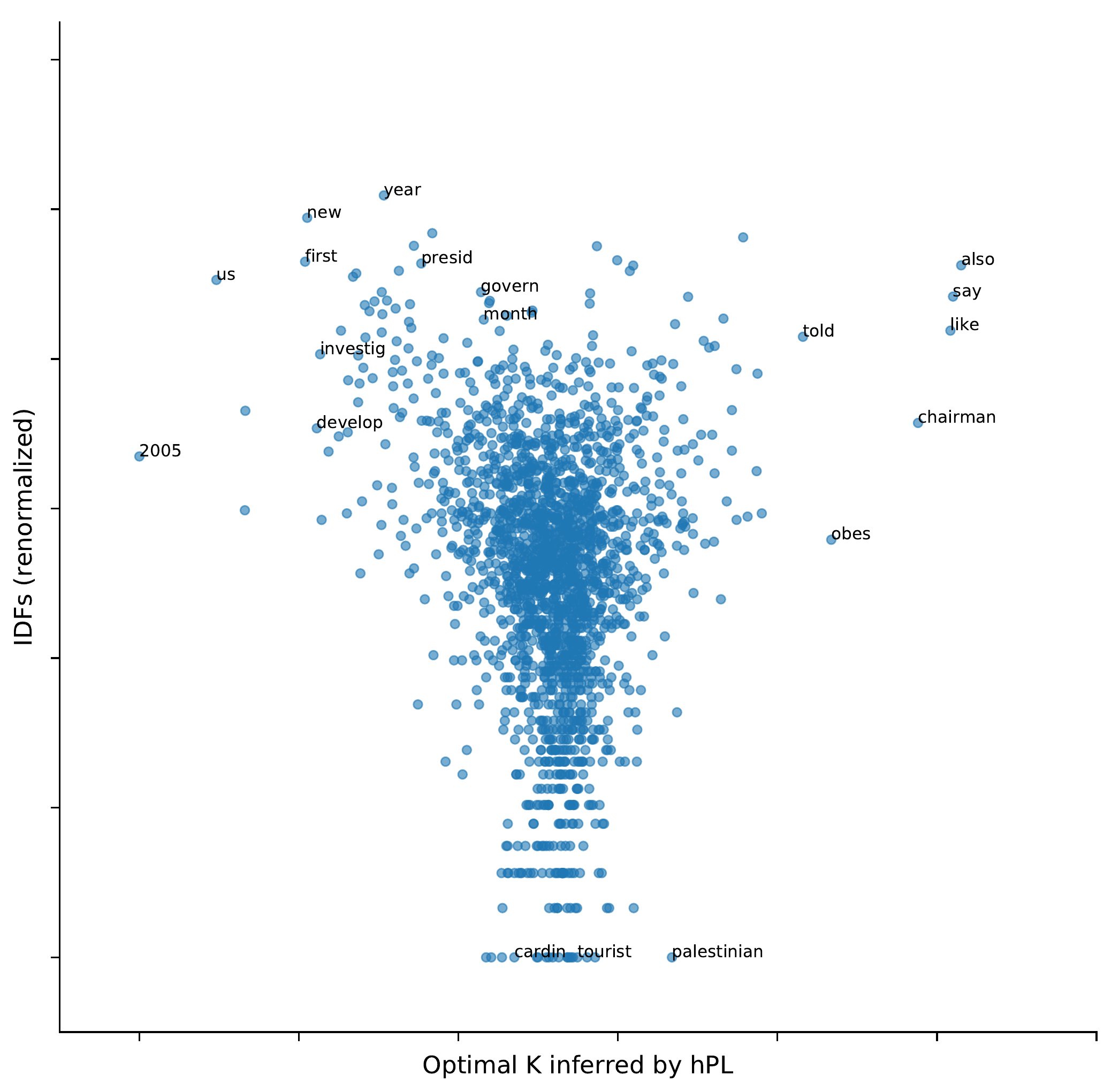}
    \caption{Scatter-plot where each dot is a word and the coordinates are its probability in $K$ and its renomalized \cpt{IDF}.}
    \label{fig:scatter_idfs_K}
    \end{subfigure}
\end{figure}

\begin{table}[]
\centering
\resizebox{0.9\columnwidth}{!}{
\begin{tabular}{@{}lcccc@{}}
\toprule
    & \multicolumn{2}{c}{TAC-08} & \multicolumn{2}{c}{TAC-09} \\
    & R-2 & MOVER & R-2 & MOVER \\
\midrule                            
\multicolumn{3}{l}{\emph{\textbf{Baselines}}} \\   
\hspace{3mm} \cpt{LR}                   &  .078 & .336 & .090 & .360  \\
\hspace{3mm} \cpt{ICSI}                 &  .101 & \textbf{.377} & .103 & .369  \\
\hspace{3mm} \cpt{KL-Greedy}            &  .074 & .294 & .069 & .289  \\
\hspace{3mm} (\cpt{JS}, Gen)            &  .101 & .375 & \textbf{.104} & .373  \\
\midrule                            
\multicolumn{3}{l}{\emph{\textbf{Ours}}} \\
\hspace{3mm} ($\theta_{U}$, Gen)      &  .098 & .353 & .094 & .359  \\
\hspace{3mm} ($\theta_{\cpt{MS|D}}$, Gen)      &  .101 & .367 & .102 & .371  \\
\hspace{3mm} ($\theta_{\cpt{hPL}}$, Gen)       &  \textbf{.104} & \textbf{.377} & .103 & \textbf{.374}  \\
\bottomrule                            
\end{tabular}
}
\caption{Comparison of summarization systems based on maximizing the summary scoring function $\theta_K$ induced by different background knowledge.}\label{tab:summary_eval}
\end{table}

\section{Comparison: IDF \vs\ optimal $K$}
\label{sec:idfs_vs_optimal}

To verify that our inferred $K$ contains different information from \cpt{ID}, we compare \cpt{IDF} and our optimal $K$ (see \Secref{sec:applications}). 

To be comparable, \cpt{IDF} weights need to be renormalized, as the \cpt{IDF} weights of known (unknown) words would be low (high) whereas $\mathbb{P}_K$ would be high (low). 
Thus, we compute $\frac{1}{C}(1-\cpt{IDF}(\omega_j))$ for each word $\omega_j$, where $C = \sum_j (1-\cpt{IDF}(\omega_j))$.

In \Figref{fig:direct_comparison}, we represent the full distributions over all words in the support of $K$ and show the absolute difference with renormalized \cpt{IDF} weights. Furthermore, \Figref{fig:scatter_idfs_K} is a scatter plot where each dot represent a word and the coordinates are its \cpt{IDF} and $K$ weights. The low correlation between the two indicates that $K$ learns a different signal than \cpt{IDF}.


        

\end{document}